\pdfoutput=1

\documentclass[11pt]{article}

\usepackage[final]{mtsummit25}

\usepackage{times}
\usepackage{latexsym}
\usepackage{enumitem} 

\usepackage[T1]{fontenc}
\usepackage{booktabs}
\usepackage[utf8]{inputenc}
\usepackage{rotating}  

\usepackage{adjustbox} 

\usepackage{CJKutf8}
\usepackage{makecell}
\usepackage{multirow}
\usepackage{microtype}

\usepackage{inconsolata}
\usepackage{amsmath} 
\usepackage{graphicx}
\usepackage{hyperref}
 \usepackage{copyright}

\title{Can \textit{Peter Pan} Survive MT? A Stylometric Study of LLMs, NMTs, and HTs in Children's Literature Translation}

\author{
 \textbf{Delu Kong\textsuperscript{1,2}}
 \textbf{ and Lieve Macken\textsuperscript{2}}
\\
\\
 \textsuperscript{1}School of Foreign Studies, Tongji University, Shanghai, 200092, China \\
 \textsuperscript{2}Language and Translation Technology Team, Ghent University, Ghent, 9000, Belgium
\\
 \small{
   \textbf{Correspondence:} \href{mailto:kongdelu2009@hotmail.com}{kongdelu2009@hotmail.com}
 }
}

\begin{document}

\begin{CJK}{UTF8}{gbsn} 

\maketitle
\begin{abstract}

This study focuses on evaluating the performance of machine translations (MTs) compared to human translations (HTs) in English-to-Chinese children's literature translation (CLT) from a stylometric perspective. The research constructs a Peter Pan corpus, comprising 21 translations: 7 human translations (HTs), 7 large language model translations (LLMs), and 7 neural machine translation outputs (NMTs). The analysis employs a generic feature set (including lexical, syntactic, readability, and n-gram features) and a creative text translation (CTT-specific) feature set, which captures repetition, rhythm, translatability, and miscellaneous levels, yielding 447 linguistic features in total.

Using classification and clustering techniques in machine learning, we conduct a stylometric analysis of these translations. Results reveal that in generic features, HTs and MTs exhibit significant differences in conjunction word distributions and the ratio of 1-word-gram-一样, while NMTs and LLMs show significant variation in descriptive words usage and adverb ratios. Regarding CTT-specific features, LLMs outperform NMTs in distribution, aligning more closely with HTs in stylistic characteristics, demonstrating the potential of LLMs in CLT.

\end{abstract}

\section{Introduction}


With the advent of LLMs\footnote{MT is used as a superordinate term encompassing both NMT and LLM translation, while NMT and LLM can also be treated as distinct categories based on the underlying generation engines.}
, various evaluations have been made within user and researcher communities~\cite{jiao_is_2023, castilho_online_2023, enis_llm_2024}. A consensus has been reached: LLMs appear to be useful for handling texts that use highly formulaic language, such as contracts, technical documents, and web pages. However, when it comes to translating highly creative texts, such as literary works, it remains a highly challenging task~\cite{kocmi_findings_2024}.

Among all text types, literary texts serve as a formidable ``bastion''~\cite{toral_is_2014} that challenge the performance of MT engines. Compared to informational texts, literary texts place greater emphasis on aesthetic creation. They are characterized by intricate linguistic structures, rich metaphors, and deep cultural nuances, interwoven with the styles of different nations, cultures, eras, and even individual authors~\cite{hadley_using_2022}. Therefore, literary translation not only facilitates the cross-linguistic transmission of ideas but also needs to recreate the artistic charm of the original work. As some scholars have pointed out, the innovative nature of literary texts ``almost implies an inherent degree of resistance to automation''~\cite[~p. 18]{ruffo_collecting_2022}, which also creates tricky challenges for MT developers seeking effective solutions.

As a significant branch of literary translation, children's literature translation (CLT for abbreviation) inspires limitless imagination in young readers worldwide. The choice to research CLT arises from its distinctive combination of literary artistry and creative expression: (1) it is crucial for cultural exchange and dissemination of children literature worldwide; (2) children's literature often features relatively unique expressions, making it ideal for assessing MT systems' semantic and cultural handling; (3) evaluating MT performance is vital to prevent mistranslations or hallucinations that could mislead young readers early on.

\section{Related work}


\subsection{Stylometric investigation in CLT studies}

Over the past decade, the field of CLT studies has expanded significantly. Scholars note that the first two decades of the 21st century have been a ``blooming period'' for research on translating children’s books~\cite{fornalczyk-lipska_repetitive_2022}. Once a marginal topic, it is now recognized that CLT works become a crucial part in young readers’ literary experiences worldwide~\cite{van_coillie_childrens_2020}. 
~\citet{puurtinen_genre-specific_2003} was among the pioneering scholars who employed a corpus-based methodology to investigate CLT. Her study examines translation-specific features in Finnish children's literature, identifying a high frequency of non-finite structures, a relative absence of colloquial expressions as potential hallmarks of translated texts.

In recent years, the study of CLT has witnessed more refined outputs within the framework of stylometry or quantitative stylistics\footnote{In this paper, stylometry is used interchangeably with quantitative stylistics, both referring to approaches that emphasize feature engineering and statistical analysis of textual style. In contrast, stylistics more broadly involves interpretive analysis of how language produces meaning, literary effects, and context-sensitive nuances.}.~\citet{cermakova_translating_2018} uses corpus-based method to analyse the feature of repetition in CLT, showing that translators often find repetition uncomfortable and tend to compensate for it by using synonymy. For the Chinese-English language pair,~\citet{zhang_explicitation_2019} examine increased lexical-grammatical explicitness in Chinese CLT, and show a higher frequency of personal pronouns compared to non-translated texts, likely due to cross-linguistic influence from the source language. Also,~\citet{zhao_translating_2022} explore how narrative space is transferred in CLT from Chinese works to English translated versions, and show that selective appropriation (patterns of omission and addition designed to suppress, accentuate or elaborate particular aspects of a narrative) is the most used strategy.

The aforementioned studies have established a foundation for applying stylometric analysis to CLT. However, compared to general translation studies, where broad features such as Type-token-ratio (TTR), PoS-tags, and word frequencies are commonly analyzed~\cite[see][]{papcke_stylometric_2022,ding_triangulating_2024,ploeger_towards_2024}, CTT-features are more specifically utilized in CLT studies and are less frequently examined in broader translation research. CLT requires attention to more specific aspects, including repetition, metaphor, and rhythmic patterns. To address this, the present study incorporates both generic stylometric features and those specifically tailored to CLT, which are presented in Section \ref{sec. featureset}.

\subsection{MT in creative texts}

MT's application in creative texts raises debates on its impact on the accuracy of cultural and creative expression. 
Even before the advent of NMT,~\citet{toral_is_2014} raised the question of the usefulness of MT for literature. The very next year,~\citet{toral_machine-assisted_2015} demonstrated that a statistical machine translation system adapted to literary texts outperformed generic baseline systems, and could be further incorporated into the literary translator’s workflow.

The NMT system advancement has sparked growing interest among scholars in applying MT to literary texts.~\citet{toral_what_2018} evaluate NMT for literary texts, specifically novels, and find that NMT achieves an 11\% relative improvement in BLEU scores compared with statistical MT. Yet if we look at literary translators’ attitudes, the story is different. Studies show more experienced literary translators prefer translating from scratch, and might be resistant in using NMTs for more freedom~\cite{moorkens_translators_2018, way_why_2023}. From a reader perspective, \citet{Guerberof_tobe_2024} observed that, in a case study of different translations, the MT version, compared to HT and PE, received the lowest ratings for narrative understanding and attentional focus, with structural and lexical issues contributing to a disrupted and confusing reading experience.

This raises the question: why are MT outputs often not well received by translators and readers? One major issue lies in stylistic limitations. \citet{daems_dutch_2022} points out that MT systems struggle with capturing stylistic nuances, humour, and contextual or cultural subtleties—elements widely recognized as central to the nature of literary texts.
~\citet{farrell_machine_2018} and~\citet{taivalkoski-shilov_ethical_2018} both emphasize that MT or post-edited output may result in homogenization and normalization in the target texts, trends that run counter to the diverse and creative nature of literary texts. Several studies investigate the stylistic difference between HTs and NMTs of literary works. For example,~\citet{jiang_corpus-based_2022} show that NMTs might be less coherent in discourse and less consistent in lexical choices. A more recent study also shows that NMT outputs can be distinguished from the HT counterparts based on sentence length distribution~\cite{rybicki_can_2025}. 
These studies inspired us to follow a similar vein and examine whether LLM outputs differ from those of NMT systems.

\subsection{Research gaps and questions}


Summarizing the above studies, despite the solid foundation of CLT research, stylometric methods remain underexplored. Additionally, while existing studies of MT on literary works predominantly focus on general literary works, children's literature has received comparatively less attention.~This gap underscores the need for research that accounts for the unique linguistic characteristics of CLT. Furthermore, the translation capabilities of LLM systems have gained significant interest; however, most research in this domain remains centered on NMT engines. The effectiveness of LLMs in CLT needs further investigation.

Based on these gaps, our study focuses on CLT among three translation groups, namely HTs, NMTs and LLMs. We construct a large dataset by incorporating a total of 21 English to Chinese translations of \textit{Peter Pan}, and establish a more comprehensive feature set with a sub-category tailored for CTT. We adopt the research design of~\citet{daems_translationese_2017} and~\citet{lynch_translators_2018}, and collect experimental results and draw salient features, and then compare inter- and intra-group performance. 

We address the following research questions:
\begin{itemize}[noitemsep, topsep=0pt]
    \item RQ1: Do MTs differ from HTs in CLT on generic textual features and CTT-specific features?
    \item RQ2: Do LLM outputs differ from NMT outputs in CLT on generic textual features and CTT-specific features?
    \item RQ3: How do salient features illustrate the differences among HTs, NMTs, and LLMs in CLT?
\end{itemize}

\section{Methodology}
\subsection{Dataset}

\begin{table*}[!h]
    \centering\small
    \begin{tabular}{l l l l l r r r l}
    \toprule
    \textbf{Type} & \textbf{Translator} & \textbf{Abbr.} & \textbf{Engine} & \textbf{Acquisition} & \textbf{Token} & \textbf{Type} & \textbf{Sent.} & \textbf{Year} \\
    \midrule
    \textbf{Source} & - & - & - & E-book & 47,978 & 5,334 & 3,334 & 2005 \\
    \midrule
    \multirow{7}{*}{\textbf{HTs}} 
    & Liang & HTL & - & OCR & 47,627 & 5,108 & 3,229 & 1929 \\
    & Yang \& Gu & HYG & - & E-book & 50,222 & 5,798 & 3,500 & 1991 \\
    & Ren & HTR & - & E-book & 51,271 & 4,968 & 3,424 & 2006 \\
    & Ma & HTM & - & E-book & 47,777 & 5,358 & 3,662 & 2011 \\
    & Sun & HSU & - & E-book & 56,674 & 5,872 & 3,673 & 2017 \\
    & Shi & HSH & - & E-book & 50,451 & 5,220 & 3,571 & 2018 \\
    & Huang & HTH & - & E-book & 51,233 & 5,477 & 3,937 & 2020 \\
    \midrule
    \multirow{7}{*}{\textbf{NMTs}} 
    & DeepL & NDL & Classic & API & 46,731 & 4,980 & 2,850 & \multirow{7}{*}{\makecell{2025\\Feb.}} \\
    & GoogleTrans & NGT & v2 & API & 46,891 & 4,887 & 3,268 & \\
    & MicrosoftTrans & NMS & - & API & 46,854 & 4,711 & 3,150 & \\
    & AmazonTrans & NAZ & - & API & 46,097 & 4,599 & 3,206 & \\
    & BaiduTrans$^{*}$ & NBD & - & API & 45,697 & 4,514 & 3,205 & \\
    & YoudaoTrans$^{*}$ & NYD & - & API & 47,817 & 4,690 & 3,393 & \\
    & NiuTrans$^{*}$ & NNT & - & API & 46,296 & 4,541 & 3,197 & \\
    \midrule
    \multirow{7}{*}{\textbf{LLMs}} 
    & ChatGPT & LCG & 4o & Web & 50,958 & 5,965 & 3,980 & \multirow{7}{*}{\makecell{2025\\Feb.}} \\
    & Claude & LCL & 3.5-sonnet & Web & 46,503 & 5,355 & 3,426 & \\
    & Gemini & LGM & 1.5-flash & API & 48,174 & 5,489 & 3,474 & \\
    & Kimi$^{*}$ & LKM & v1 & API & 57,320 & 5,221 & 3,869 & \\
    & DeepSeek$^{*}$ & LDS & v3 & OpenSource & 45,951 & 4,865 & 3,421 & \\
    & TowerInstruct & LTI & 7b-v0.2 & OpenSource & 46,097 & 4,869 & 3,124 & \\
    & LaraTrans & LLT & Creative & Web & 48,576 & 4,709 & 3,176 & \\
    \midrule
    \textbf{Total} & - & - & - & - & \textbf{1,025,217} & \textbf{107,196} & \textbf{71,735} & - \\
    \bottomrule
    \end{tabular}
\caption{Overview of the datasets used in this study. Pure NMT version is selected if the provider offers version options (such as NDL and NGT). LaraTranslate provides a ``creative text'' style option but, despite claiming to use LLM technology, does not support custom prompts. Engines marked with $^{*}$ are developed by Chinese enterprises. The total count excludes the source text.} 
\label{tab: dataset}
\end{table*}

The corpus used in this study is based on J.M. Barrie’s 1911 novel \textit{Peter and Wendy}, the classic children's version of the Peter Pan story\footnote{The 1911 novel uniquely includes the final chapter ``When Wendy Grew Up'', which does not appear in play or later abridged versions. This chapter is present in all Chinese translations used in this study. Although some translations do not explicitly state the source text, the inclusion of this chapter indicates that all were ultimately based on the 1911 novel.}. The character of Peter Pan was first introduced in Barrie’s 1902 novel \textit{The Little White Bird}. The 1911 novel has been widely cherished by children worldwide for its vivid imagination and the thrilling adventures of Peter Pan. The HTs are selected from \textbf{7} Chinese translations published by reputable publishers, spanning a wide time range (1929–2020) to ensure representativeness across different eras.

For MTs, the translated texts are categorized into two groups: NMTs and LLMs, with \textbf{7} engines selected for each category to produce a total of \textbf{14} MTed versions. In the NMT category, while some providers have reportedly begun integrating LLM technologies into their traditional NMT systems, e.g. DeepL\footnote{\url{https://www.smartling.com/blog/how-accurate-is-deepl}}, this study specifically aims to examine ``pure'' NMTs. To achieve this goal, models that explicitly state using an NMT engine are prioritized. For providers offering multiple MT engine versions, the NMT variant is selected whenever possible. If no version details were disclosed, the default engine is used. All NMTs are accessed via API. Details can be viewed in the online supplementary materials. 

For LLMs, we select state-of-the-art commercial generic models, such as ChatGPT and Claude, as well as open-sourced DeepSeek, and MT-tailored LaraTranslate, and both open-sourced and MT-tailored Unbabel-TowerInstruct. Large-parameter deep reasoning models such as GPT-o1 and DeepSeek-R1, though strong in contextless multilingual translation, are not adopted in our framework due to their high inference cost, slower processing speed, and tendency to generate rambling outputs in Chinese~\cite{chen2025evaluatingo1likellmsunlocking}. These factors significantly increase computational complexity and reduce overall efficiency in translation tasks. Consequently, we prioritize LLM models optimized for faster, more direct translation processes.

The MT process for LLMs involves prompt engineering, adhering to practices outlined in Andrew Ng’s course\footnote{\url{https://learn.deeplearning.ai/courses/chatgpt-prompt-eng}} and the CRISPE framework\footnote{\url{https://github.com/mattnigh/ChatGPT3-Free-Prompt-List}}. This ensures a structured and standardized prompt design, consistently applied across all engines during translation. The complete prompt used in our experiments is provided in Appendix \ref{sec:appendix a} for reference. Furthermore, we distinguish between MT engines developed by Chinese and non-Chinese enterprises, as this distinction may influence translation strategies and quality, particularly given that Chinese is the target language.

All texts underwent rigorous preprocessing, including cleaning, denoising, part-of-speech (PoS) tagging, and dependency (Dep) parsing. Since Chinese lacks explicit word boundaries, prior word segmentation is necessary. To achieve SOTA performance, we utilize the Language Technology Platform (LTP)\footnote{\url{https://github.com/HIT-SCIR/ltp}}, a comprehensive natural language processing toolkit~\cite{che_n-ltp_2021}. LTP’s deep learning model (Base2) is used for word segmentation, PoS tagging, and syntactic analysis, achieving reported accuracies of 99.18\%, 98.69\%, and 90.19\% for these tasks, respectively.\footnote{\url{https://github.com/HIT-SCIR/ltp/blob/main/README.md}}

The final \textit{Peter Pan} translation corpus exceeds over one million tokens. A detailed overview of the dataset is provided in Table \ref{tab: dataset}.

\subsection{Feature set}
\label{sec. featureset}

Based on the principles to construct feature set for stylometric analysis~\cite{volansky_features_2013}\footnote{Although the quoted research focuses on translationese study, their proposed principles are highly relevant to construct a well-balanced and interpretable feature set for studying linguistic features of translated texts.} and referring to previous research~\cite[see][]{huang_wei_application_2009, lynch_translators_2018, toral_post-editese_2019,de_clercq_uncovering_2021}, the following section presents the feature set applied in this study. A brief feature summary is in Table~\ref{tab:features} (in Appendix \ref{sec:appendix b}). All together, we've employed 447 features in this study. It should be noted that all features are represented as ratios or weighted measures to mitigate the influence of sample size differences and ensure comparability across texts.

\subsubsection{Generic textual features}

The generic textual features are designed to capture text characteristics from a holistic perspective. ``Generic'' means that these features are commonly used in other types of stylometric studies, not constrained on CLT studies. They are divided into four linguistic dimensions: lexical, syntactical, readability, and N-gram features. 

Lexical features reflect word-level characteristics on lexical diversity, density, and richness, such as TTR. Additionally, PoS tag features, extracted using the LTP platform\footnote{\url{https://ltp.ai/docs/appendix.html\#id2}}, capture the distribution of word classes, such as nouns and verbs.

Syntactic features focus on overall sentence structure and syntactic patterns, including average sentence length. Dependency tags are used to further analyze syntactic roles, such as Mean Dependency Distance and the ratio of head and node.

Readability features include nine readability metrics proposed by~\citet{lei_alphareadabilitychinese_2024}\footnote{\url{https://github.com/leileibama/AlphaReadabilityChinese}}, which evaluate lexical, syntactic, and semantic variability to assess a text’s difficulty and comprehensibility for the target audience. Moreover, four concreteness features are included, measuring lexical concreteness based on the work of~\citet{xu_concretenessabstractness_2020}.

N-gram features utilize N-word-grams and N-PoS-grams, where N ranges from 1 to 3, to capture locally constrained phrasal patterns. These features are extracted by comparing the target corpus with the LCMC reference corpus\footnote{\url{https://www.lancaster.ac.uk/fass/projects/corpus/LCMC/}}. To maintain consistency, LCMC was re-tagged using the same LTP tools to ensure a uniform PoS-tag set.

\subsubsection{CTT-specific features}

CTT-specific features refer to characteristics specifically designed for creative text translation. In this study, we incorporate insights from previous Chinese research on CLT. These features are tailored to the linguistic and stylistic elements of CLT and are categorized into four subcategories:

Repetition features are a widely recognized in CLT, as a narrative strategy, often used to capture young readers' attention~\cite{mastropierro_avoidance_2022}. In this study, repetition features include AA-pattern (two-character repetition), AAA-pattern (three-character repetition), and ABAB-pattern structures. 

Rhythm features examine phonetic patterns within the text, as previous research has identified distinct rhythmic patterns in CLT~\cite{cooper_rhythm_1989}. This study captures key rhythm features, such as rhyme proportion, vowel balance, and tonal alternation.

Translatibility features\footnote{We categorize translatability features under the CTT-specific level, considering that CLT is particularly sensitive to language shifts. This sensitivity is especially pronounced when the target audience is mainly children. The interweaving of English elements within Chinese text might influence their reading experience.} assess language transfer and translation completeness between the source and the target texts through five key aspects: completeness, foreignness, code-switching, abbreviation, and untranslatable elements. Notably, completeness identifies untranslated English phrases longer than three words, while foreignness measures the ratio of English to Chinese characters.

The miscellaneous category includes five unique linguistic features that could not satisfactorily fall into the previous classifications but are frequently observed in CLT. These include the proportion of onomatopoeia, the usage of the Chinese-specific ``-er suffix''\footnote{The ``-er suffix'' in Chinese is a linguistic phenomenon where the suffix "儿" (ér) is added to the end of a word, often altering its pronunciation and adding a diminutive meaning. This is widely used in spoken and literary Chinese. For example, 花 (huā, "flower") becomes 花儿 (huār), and 鸟 (niǎo, "bird") becomes 鸟儿 (niǎor).}, and the frequency of sentence-final particles, among others.

We put more detailed explanations and examples in the supplementary materials.

\subsection{Algorithms}

\subsubsection{Feature selection}

To streamline the experiment, mitigate feature noise, and enhance efficiency, we adopt a feature selection strategy based on chi-square (\(\chi^2\)) ranking in both classification and clustering tasks.
Features are prioritized according to their \(\chi^2\) values, with the top 30 being retained. If a given category contains fewer than 30 features, all available ones are preserved.

\subsubsection{Classification experiment}

The classification experiment follows a hierarchical structure based on different feature set levels. Initially, classification is performed separately for each feature sub-level. The experimental setup consists of the following comparison groups: (1) HTs vs. MTs, where MTs encompass both NMTs and LLMs; (2) HTs vs. NMTs and LLMs separately; (3) LLMs vs. NMTs; and (4) intra-group classification within the NMTs and LLMs categories.

To evaluate classification performance, five classifiers are employed: Naïve Bayes, Logistic Regression, Support Vector Machine (SVM), Decision Tree, and Random Forest. The average performance across these classifiers is reported. The SVM model is configured with a linear kernel, while the remaining classifiers use their default settings. Following the methodology outlined by~\citet{rahman_commentclass_2024}, the effectiveness of the ensemble classifier is assessed using Accuracy (ACC) score. All classification tasks, except intra-group classifications within the NMTs and LLMs groups, are binary classification tasks.

\subsubsection{Clustering experiment}

The clustering experiment employs the \(k\)-means algorithm to categorize data. Rather than predefining the number of clusters (\(k\)), it is determined based on performance evaluation.~Euclidean distance serves as the similarity metric for clustering, and the feature set is derived from the Top-k features identified in the earlier analysis.

To assess clustering effectiveness, the Adjusted Rand Index (ARI) is used as the primary evaluation metric. ARI quantifies the alignment between the clustering results and ground truth labels while accounting for chance, providing an objective measure of clustering quality~\cite{warrens_understanding_2022}. Beyond numerical evaluation, interactive clustering visualizations are generated using Python’s Plotly library.

In addition to \(k\)-means clustering, hierarchical clustering is introduced as a complementary approach. This method utilizes the stylo package in R, incorporating the top 100 most frequent words and Eder's delta~\cite{eder_rolling_2015} as key metrics. By integrating multiple clustering strategies, we aim to strengthen the robustness of our analysis and enhance the overall reliability of the experimental findings.

\section{Results}
\subsection{Classification}

\begin{table}[h]
\centering
\small
\renewcommand{\arraystretch}{1.2}
\setlength{\tabcolsep}{3.5pt} 

\resizebox{\linewidth}{!}{ 
\begin{tabular}{llcccc}
\toprule
\textbf{Level} & \textbf{Sub-level} & \textbf{\makecell{HTs\\-MTs}} & \textbf{\makecell{HTs\\-NMTs}} & \textbf{\makecell{HTs\\-LLMs}} & \textbf{\makecell{NMTs\\-LLMs}} \\
\midrule
\multirow{4}{*}{\makecell{Generic\\textual\\features}} 
& Lexical         & 0.8673 & 0.9149 & 0.8526 & 0.7439 \\
& Syntactical     & 0.8001 & 0.8443 & 0.7128 & 0.6752 \\
& Readability     & 0.6645 & 0.7056 & 0.5538 & 0.6178 \\
& N-gram          & 0.8674 & 0.8803 & 0.8486 & 0.7724 \\
\midrule
\multirow{4}{*}{\makecell{CTT-\\specific\\features}} 
& Repetition      & 0.6274 & 0.6800 & 0.5778 & 0.5436 \\
& Rhythm          & 0.6650 & 0.5564 & 0.6444 & 0.5593 \\
& Translatibility & 0.6883 & 0.7930 & 0.6091 & 0.6067 \\
& Miscellaneous   & 0.7378 & 0.7521 & 0.6393 & 0.6172 \\
\midrule
All & - & 0.9149 & 0.9376 & 0.8896 & 0.7464 \\
\bottomrule
\end{tabular}
}
\caption{Classification results across different feature levels and comparison groups. ``Generic textual features'' encompass general linguistic attributes, while ``CTT-specific features'' focus on aspects relevant to creative text translation. ``All'' represents the combined performance when all features are used together.}
\label{tab: class}
\end{table}

Table~\ref{tab: class} presents the results across feature levels and groups. It should be noted that a high ACC score indicates a clear distinction between the examined categories, while a low ACC score suggests greater similarity.
We generalize two tendencies: 

First, from group comparison perspective, HTs consistently achieve the highest performance across different pairwise comparisons, with HTs-NMTs reaching the highest accuracy (0.9376) and HTs-MTs following closely (0.9149). When LLMs are involved, accuracy drops, as seen in HTs-LLMs (0.8896) and NMTs-LLMs (0.7464). This suggests that distinguishing between HTs and other MTed translations (NMTs and LLMs) is relatively easier, while differentiating within the same category (intra-group) is much harder, as in HTs (0.6785), NMTs (0.5965), and LLMs (0.5917).

Second, from feature categories perspective, among the generic textual features, lexical and N-gram features contribute the most to classification, with the highest ACC across different translation types (e.g., HTs-NMTs: 0.9149 and 0.8803, respectively). Readability exhibits the lowest performance, as in HTs-LLMs (0.5538). Regarding CTT-specific features, translatability and miscellaneous features contribute notably in differentiating HTs from other groups, while repetition and rhythm drop sharply in ACC scores. Overall, generic features perform better than CTT-specific features in classification, and using all features leads to the best performance.

Figure~\ref{fig:heatmap} (in Appendix \ref{sec:appendix d}) presents a pairwise classification heatmap to provide a visualized plot and a fine-grained classifying result. It reveals three main results:

First, for HTs, the classification accuracy between HTs and other groups (NMTs and LLMs) is generally high. Within the HTs group, the highest ACC exceeds 0.90. The highest intra-group accuracy is observed in the HTL-HTH pair (0.98), also HTL achieves the highest average ACC (0.97) compared with all other samples. This may be due to the significant temporal gap, as HTL is the earliest translation among HTs.
Conversely, the HSU-HTM and HSU-HTR pairs (both 0.76) exhibit the lowest accuracy, with HSU having the lowest average ACC (0.87). The greater similarities among these translations might be newer versions drawing references from previously published ones.

Second, for NMTs, ACC in distinguishing NMTs-LLMs is relatively lower than that of NMTs-HTs, meaning NMTs are much more similar to LLMs in style. The lowest inter-group ACC is observed in NGT-LDS (0.59). The score within the NMTs group is considerably lower than that within the HTs group. The highest intra-group accuracy is NAZ-NYD (0.94), while the lowest is NBD-NGT (0.44). Also, in terms of average ACC, we found that NAZ (0.92) and NDL (0.9) achieve relatively higher compared with the rest NMT engines, while NGT (0.79) is the lowest.

Third, within the LLM group, the highest intra-group ACC is observed in LCG-LLT (0.99), and LKM-LLT (0.65) exhibits the lowest. Still, on average ACC, LCG (0.96) and LCL (0.93) have the relative highest score compared with other LLM engines, while LDS (0.83) has the lowest one. No significant differences are found between MT systems developed by Chinese companies and those by international companies.

\subsection{Clustering}

Figure~\ref{fig:cluster} presents the clustering results using K-means (left) and hierarchical clustering (right). The K-means clustering, with an ARI score of 0.4873, demonstrates a clear separation between HTs and NMTs, as the two groups are positioned far apart. However, LLMs exhibit a more complex distribution, which cannot be clustered into a distinct group. Notably, three LLMs (LCG, LCL, LGM) cluster closer to HTs, suggesting that their translations share more similarities with human translations. Meanwhile, a subset of LLMs (LKM, LDS, LLT, LTI) aligns more closely with NMTs. 

The hierarchical clustering (right) supports these findings, displaying relatively stable and well-separated clusters for HTs and NMTs. In contrast, LLMs show a more dispersed pattern, with samples integrating into both HT and NMT clusters. This reveals that LLMs exhibit heterogeneous translation characteristics, with some models leaning towards human-like translation styles and others resembling NMT outputs. Both clustering results reinforce observations drawn from previous classification experiments.

\section{Discussion}

\subsection{Overview from generic features}


This section discusses these differences from the perspective of generic features and provides an overview on their variance. 

\subsubsection{Ratio of conjunctions}

From Table~\ref{tab:salient_features}, we can see that conjunction words and N-grams contribute greatly to the separation of HTs and MTs. For the ratio of conjunction words (Figure \ref{fig: anova_conj}), HTs differ significantly from MTs, with MTs exhibiting a higher ratio (ANOVA F = 91.10, \textit{p} < 0.0001)\footnote{To determine significant differences, we first conduct a normality test on the data. If the data met the normality assumption, we apply ANOVA; otherwise, we use the non-parametric Kruskal-Wallis test.}. Among conjunction words, the 1-word-gram-然后 (English: then) serves as a good example, since it is particularly prominent, and exhibits a similar trend on its over-usage in MT outputs. It means that MTs tend to rely more on explicit logical connectors, exhibiting a certain tendency toward ``explicitation''~\cite{zhang_explicitation_2019}, whereas HTs demonstrate greater flexibility in expression and are not strictly bound by the logical transitions of the source text. However, no significant difference is observed between NMTs and LLMs in terms of conjunction (p = 0.06). 

\subsubsection{Ratio of 1-word-gram-一样}

Another feature that distinguishes HTs and MTs apart is the 1-word-gram-一样 (same). This word frequently follows another word to form a Chinese phrase ``像...一样 (same as...)'' which conveys a simile meaning. As shown in Figure \ref{fig: anova_sameas}, MTs exhibit a significantly higher frequency of ``一样'' compared to HTs (p < 0.0001), indicating that MTs tend to produce more explicit comparative structures. Particularly NMTs use ``像...一样'' more frequently. Hence, MTs are more constrained by source text structures and produce similar patterns, leading to potential `homogenisation'~\cite{daems_impact_2024} in lexical terms with less variation in figurative expressions. 
But for LLMs, we see that LCG (ChatGPT) and LCL (Claude) use relatively less ``像...一样''. Given that the principle of fidelity to source texts should be generally maintained in translation, the reduction in ``像...一样'' likely reflects a shift in how figurative meaning is expressed in the LLM translations, rather than a loss of figurative content.


Drawing on actual concordance as an example (see Figure~\ref{fig: Concordance_sameas}), we observe that in HTs, there are only two occurrences of ``像...一样'', while other variations, such as ``像...似的'' and ``宛如'', are also used creatively. In contrast, all seven NMT outputs employ the same fixed expression, whereas LLMs exhibit a mix, with three outputs using the same phrase.

\subsubsection{Ratio of descriptive and adverbial words}

Figure~\ref{fig: adverb} illustrates two important features that show significant differences between NMTs and LLMs. For the proportion of descriptive words (ratio\_dscrptW), texts translated by LLMs exhibit a significantly higher usage of descriptive expressions than NMTs (p < 0.0001). Since descriptive words are generally regarded as enhancing textual vividness and specificity by providing richer contextual details, their higher occurrence in LLMs suggests more expressive and stylistically nuanced outputs than NMTs. Second, the proportion of adverbs (ratio\_adverb) across HTs, NMT, and LLM indicates significant differences among these three systems (\textit{p} < 0.0001). The trend suggests that HTs employ adverbs more frequently than MTs, where LLM restores some adverb usage compared to NMT.

\subsection{Zooming into CTT-specifc features}

If we narrow down our analysis to a more specific CLT perspective, we can see from Table \ref{tab:salient_features} that several CTT-specific features also stand out in distinguishing different translation groups. 

\subsubsection{Ratio of foreignness}

To begin with, the left column of Figure \ref{fig: anova_combined} presents the foreignness feature at the level of translatability. By definition, this feature quantifies the ratio of English words that appeared in the translated text. HTs exhibit zero occurrence of the retained foreign words. In contrast, MTs, particularly NMTs, demonstrate a notably higher ratio. Interestingly, LLMs display a substantially lower foreignness ratio compared to NMT, approaching HT-like tendencies. 

Most untranslated cases are names (Such as ``其次是Slightly'' in NBD; ``Tink确实又开始四处乱窜'' in NGT), and idiomatic expressions (``他与他们分道扬way了'' in NAZ). Although this is less common in LLMs, some expressions are still translated incompletely (``他们 perfectly safe，不是吗？'' in LGM), a type of error also pointed out by~\citet{macken_machine_2024}. For child readers, minimizing source-language element leakage in CLT is a way both to improve acceptability and to mitigate ``cultural colonialism'' , since children's limited experience may necessitate a higher degree of adaptation than adult fiction~\cite[p.~38]{lathey_translating_2015}. In this regard, LLMs demonstrate an advantage over NMTs.

\subsubsection{Ratio of er-suffix}

For the ``ratio\_er\_suffix'' feature, as shown in the middle column of Figure~\ref{fig: anova_combined}, the upper graph indicates that HTs employ significantly more ``-er suffixes'' than MTs, while in the lower graph, LLMs exhibit a higher ratio compared to NMTs. 

The ``-er suffix'' serves as an important figurative expression in CLT, as it often conveys a colloquial, playful, or affectionate tone. It is commonly used in northern Chinese dialects, particularly in Beijing~\cite[see][]{chen_tone_2000,fu_contrast_2022}. Such phonetic modifications cater to children's cognitive development and linguistic preferences, making texts more engaging and accessible. Based on this, HTs remain the most effective in preserving ``-er suffix'' in CLT. However, prompt-tuned LLMs demonstrate a stronger ability to capture ``-er suffix'' features compared to NMTs, suggesting that LLMs are more aligned with HTs in this aspect.

\subsubsection{Ratio of repetitive expression}

Repetitive expressions, such as AA (e.g., ``热热的'' warm and cozy), AAA (e.g., ``慢慢慢'' very slowly), and ABAB (e.g., ``很久很久'' a very long period), are a prominent stylistic feature in children's literature. The ratio\_AA\footnote{It should be noted that the AA pattern mentioned here excludes fixed proper nouns, such as ``妈妈 mom''.} feature in the right column of Figure~\ref{fig: anova_combined} indicates that HTs employ significantly more repetition than MTs, while LLMs outperform NMTs in preserving this pattern. Research has shown that repetitive structures enhance readability, reinforce linguistic patterns, and facilitate memory retention for young readers, making texts more engaging and accessible~\cite{tannen_talking_1989, hickmann_childrens_2003}. Given these, the results suggest that LLMs better capture the stylistic and cognitive functions of repetition in CLT than NMTs, making them more aligned with HTs.

\subsection{Some further remarks on the LLM translations}

\subsubsection{On LCG and LCL}

Figure~\ref{fig:heatmap} shows that two LLM-based engines, LCG (ChatGPT) and LCL (Claude) exhibit significant distinctions in pair-wise classification compared to HTs and other MTs. By examining the feature importance list in the classification logs, we observe that LCG demonstrates exceptionally high divergence in the \textit{Average Number of Children per Node} feature (see Figure \ref{fig:llm}), exceeding the values of other texts by approximately 1.5 times. Specifically, HT averages around 20, while LCG reaches approximately 35.

This feature reflects the syntactic dependency tree structure, where a higher value suggests LCG favors a flatter syntactic structure rather than a deeply nested one. LCG appears to prioritize parallelism and broader phrase expansion. Corpus analysis (see online supplementary material) further corroborates this pattern where LCG tends to segment sentences more frequently, breaking complex structures into multiple shorter clauses deliberately, perhaps in order to retain readability suitable to children's levels.

Subsequently, we observed that LCL exhibits a significantly higher deviation in the ratio\_quote (quotation mark) feature, reaching approximately twice the value of other engines (see Figure~\ref{fig:llm}). While other LLMs maintain an average ratio of 0.13, LCL reaches approximately 0.25. Although \textit{Peter Pan} is a children's novel rich in dialogue, such an unusually high occurrence of quotation marks appears atypical.

Upon inspecting the corpus, we found that other LLMs use directional Chinese quotation marks, whereas LCL employs non-directional English quotation marks. Our quotation-matching process was designed to recognize left quotation mark and non-directional mark, but not for right quotation mark. This explains why LCL's quotation mark count is nearly double that of other models. However, in formal writing, Chinese translations should adhere to standard typographic conventions, using directional Chinese quotation marks. In this regard, LCL's handling of punctuation is less consistent with formal Chinese writing norms compared to other engines.

\subsubsection{On LDS, LTI and LLT}

In this section, we discuss in more detail the recent open-source engine LDS (DeepSeek) and MT-tailored engine LLT (LaraTranslate), and both MT-tailored and open-sourced engine LTI (Unbabel Tower\_instruct). 

The three engines exhibit a high degree of confusion with other MT engines in classification tasks, with LDS and NGT achieving classification accuracies of only 0.59 and 0.66 with NBD, respectively. While these engines (particularly LDS) have gained significant attention recently, a more critical evaluation reveals that their performance in MT tasks shows no substantial improvement compared to other LLM-based engines.

Among MT-tailored LLMs, LLT and LTI fail to reach a satisfactory level in CTT-specific features with notably poorer performance than other commercial LLM models (see Figure~\ref{fig:llm}). Specifically, they exhibit lower usage of the ``-er suffix'' and fewer repetitive expressions. Despite LLT’s popularity and its advertised ``creative translation" capabilities, our analysis finds little evidence of enhanced stylistic performance in its output. Additionally, during our testing, LLT produced a significant number of hallucinations, and its trans\_foreignness score was the highest.

Some marketing claims about certain LLMs should be critically evaluated rather than taken at face value. While some models may perform well in general machine translation tasks, their effectiveness in specialized domains—such as children's literature—requires thorough empirical validation.

\section{Conclusion}

This study investigates the extent to which MTs diverge from HTs in CLT from a stylometric perspective, focusing on both generic textual features and CTT-specific features.

For RQ1, our findings confirm that MTs exhibit significant differences from HTs across both feature sets. Looking at generic text features, MTs deviate from HTs in conjunction word distributions and the ratio of 1-word-grams, where MTs tend to favor literal translation strategies, with a stronger influence from the source text. Looking at CTT-specific features, MTs generally fail to reproduce stylistic elements crucial in CLT, such as repetition and ``-er suffix'', further reiterating the challenges of automated translation in preserving literary expressiveness.

For RQ2, NMTs and LLMs diverge notably in both generic and CTT-specific stylistic features. While LLMs and NMTs exhibit significant differences in descriptive word usage and adverb ratios, LLMs show greater alignment with HTs.
LLMs also outperform NMTs in ``-er suffix'' usage, AA-pattern repetition, and foreignness. Thus, they are better and more effective at capturing stylistic patterns in CLT than NMTs.

For RQ3, our analysis reveals distinct stylistic differences among HTs, NMTs, and LLMs. HTs still act as the gold standard in stylistic expressions, while NMTs produce more rigid and less engaging outputs. LLMs strike a balance, with greater stylistic fluidity than NMTs and approximating HT-like translation patterns. However, performance varies across LLMs. ChatGPT and Claude exhibit stronger stylistic consistency, whereas open-sourced or MT-tailored models show no clear advantages over the rest.


\section*{Limitations and future work}
This study has several limitations that future research should address:

First, in terms of dataset selection, this study is limited to a single literary work, \textit{Peter Pan}, which, while representative, may not fully capture the diversity of children's literature.~Future research should expand the dataset to include translations of varied genres and styles to improve generalizability. Moreover, it should be noted that the training data of LLMs may contain human translations of \textit{Peter Pan}, potentially influencing the results and blurring the boundaries between human and LLM translations.

Second, methodologically, this study primarily relies on quantitative stylometric analysis, offering a broad ``distant reading'' of translation patterns. However, it lacks in-depth qualitative analysis to explain why certain stylistic deviations occur between HTs, LLM and NMT outputs. Future work could incorporate qualitative case studies or human evaluations to better understand how MT outputs impact the reading experience of child audiences and whether stylistic deficiencies could be mitigated through prompt engineering or fine-tuning techniques.

Lastly, the feature set in this study remains lexical and syntactic-centric, with limited exploration of semantic and discourse-level attributes. Some feature overlaps were also observed, which could introduce redundancy in classification tasks. Future work should incorporate feature correlation analysis and dimensionality reduction methods (e.g., PCA) to refine the feature set and explore network-based approaches for a more holistic view of stylistic variations.

\section*{Acknowledgments}
We gratefully acknowledge support from the China Scholarship Committee for the Visiting PhD Project (Num.~202406260211). We also thank the three anonymous reviewers for their constructive feedback.

\section*{Supplementary material}
The supplementary material used in this study is uploaded onto Github \url{https://github.com/DanielKong1996/CLT_MTsummit}

\section*{Sustainability statement}
This study primarily utilizes commercial MT engines' APIs during the translation acquisition phase. Due to the nature of these proprietary systems, accurately estimating the associated carbon footprint is challenging. Additionally, the classification and clustering experiments conducted during the machine learning phase of this research require relatively low computational resources. All experiments were performed on a personal laptop, ensuring minimal energy consumption. As a result, the overall environmental impact of this research is expected to be low.

\bibliography{references, mtsummit25}

\begin{thebibliography}{48}
\providecommand{\natexlab}[1]{#1}

\bibitem[{Castilho et~al.(2023)Castilho, Mallon, Meister, and Yue}]{castilho_online_2023}
Sheila Castilho, Clodagh Mallon, Rahel Meister, and Shengya Yue. 2023.
\newblock \href {https://aclanthology.org/2023.eamt-1.39} {Do online machine translation systems care for context? {What} about a {GPT} model?}
\newblock Tampere, Finland. European Association for Machine Translation (EAMT).

\bibitem[{Che et~al.(2021)Che, Feng, Qin, and Liu}]{che_n-ltp_2021}
Wanxiang Che, Yunlong Feng, Libo Qin, and Ting Liu. 2021.
\newblock \href {https://doi.org/10.18653/v1/2021.emnlp-demo.6} {N-{LTP}: an open-source neural language technology platform for {Chinese}}.
\newblock In \emph{Proceedings of the 2021 {Conference} on {Empirical} {Methods} in {Natural} {Language} {Processing}: {System} {Demonstrations}}, pages 42--49, Online and Punta Cana, Dominican Republic. Association for Computational Linguistics.

\bibitem[{Chen et~al.(2025)Chen, Song, Zhu, Chen, Yang, Zhao, and zhang}]{chen2025evaluatingo1likellmsunlocking}
Andong Chen, Yuchen Song, Wenxin Zhu, Kehai Chen, Muyun Yang, Tiejun Zhao, and Min zhang. 2025.
\newblock \href {https://arxiv.org/abs/2502.11544} {Evaluating o1-like llms: Unlocking reasoning for translation through comprehensive analysis}.
\newblock \emph{Preprint}, arXiv:2502.11544.

\bibitem[{Chen(2000)}]{chen_tone_2000}
Matthew~Y. Chen. 2000.
\newblock \href {https://books.google.com/books?hl=zh-CN&lr=&id=D328u70WNgMC&oi=fnd&pg=PP1&dq=Tone+Sandhi+Patterns+across+Chinese+Dialects&ots=WZu_cOspvU&sig=QECPyj-J6hzT1ZfEYt5Z4AfcFV0} {\emph{Tone sandhi: patterns across {Chinese} dialects}}, volume~92.
\newblock Cambridge University Press.

\bibitem[{Cooper(1989)}]{cooper_rhythm_1989}
B.~Lee Cooper. 1989.
\newblock \href {https://doi.org/10.1080/03007768908591344} {Rhythm ‘n’ rhymes: character and theme images from children's literature in contemporary recordings, 1950–1985}.
\newblock \emph{Popular Music \& Society}.

\bibitem[{Daems(2022)}]{daems_dutch_2022}
Joke Daems. 2022.
\newblock \href {https://library.oapen.org/handle/20.500.12657/59125} {Dutch literary translators' use and perceived usefulness of technology: the role of awareness and attitude}.
\newblock In \emph{Using {Technologies} for {Creative}-text {Translation}}. Routledge.

\bibitem[{Daems et~al.(2017)Daems, De~Clercq, and Macken}]{daems_translationese_2017}
Joke Daems, Orphée De~Clercq, and Lieve Macken. 2017.
\newblock \href {https://doi.org/10.52034/lanstts.v16i0.434} {Translationese and post-editese: how comparable is comparable quality?}
\newblock \emph{Linguistica Antverpiensia, New Series – Themes in Translation Studies}, 16:89--103.

\bibitem[{Daems et~al.(2024)Daems, Ruffo, and Macken}]{daems_impact_2024}
Joke Daems, Paola Ruffo, and Lieve Macken. 2024.
\newblock \href {https://aclanthology.org/2024.ctt-1.6/} {Impact of translation workflows with and without {MT} on textual characteristics in literary translation}.
\newblock In \emph{Proceedings of the 1st {Workshop} on {Creative}-text {Translation} and {Technology}}, pages 57--64, Sheffield, United Kingdom. European Association for Machine Translation.

\bibitem[{De~Clercq et~al.(2021)De~Clercq, De~Sutter, Loock, Cappelle, and Plevoets}]{de_clercq_uncovering_2021}
Orphée De~Clercq, Gert De~Sutter, Rudy Loock, Bert Cappelle, and Koen Plevoets. 2021.
\newblock \href {https://hal.science/hal-03406287/} {Uncovering machine translationese using corpus analysis techniques to distinguish between original and machine-translated {French}}.
\newblock \emph{Translation Quarterly}, (101):21--45.

\bibitem[{Ding(2024)}]{ding_triangulating_2024}
Guoqi Ding. 2024.
\newblock \href {https://doi.org/10.1093/llc/fqae039} {Triangulating text relationship in literary retranslation: {The} {Great} {Gatsby} in {Chinese}}.
\newblock \emph{Digital Scholarship in the Humanities}, (39):849--863.

\bibitem[{Eder(2015)}]{eder_rolling_2015}
Maciej Eder. 2015.
\newblock \href {https://doi.org/10.1093/llc/fqv010} {Rolling stylometry}.
\newblock \emph{Digital Scholarship in the Humanities}, 31(3):457--469.

\bibitem[{Enis and Hopkins(2024)}]{enis_llm_2024}
Maxim Enis and Mark Hopkins. 2024.
\newblock \href {https://doi.org/10.48550/arXiv.2404.13813} {From {LLM} to {NMT}: advancing low-resource machine translation with {Claude}}.
\newblock \emph{arXiv preprint}.

\bibitem[{Farrell(2018)}]{farrell_machine_2018}
Michael Farrell. 2018.
\newblock \href {https://apeiron.iulm.it/handle/10808/47325} {Machine translation markers in post-edited machine translation output}.
\newblock In \emph{Proceedings of the 40th {Conference} {Translating} and the {Computer}}, pages 50--59.

\bibitem[{Fornalczyk-Lipska(2022)}]{fornalczyk-lipska_repetitive_2022}
Anna Fornalczyk-Lipska. 2022.
\newblock \href {https://doi.org/10.24193/JRHE.2022.2.2} {Repetitive or innovative? {Children}’s literature in translation as the main focus of {B}.{A}. and {M}.{A}. theses}.
\newblock \emph{Journal of Research in Higher Education}, 6:38--51.

\bibitem[{Fu(2022)}]{fu_contrast_2022}
Boer Fu. 2022.
\newblock \href {https://doi.org/10.16995/glossa.5801} {Contrast preservation in mandarin {R}-suffixation: a comparative study of beijing and liaoning dialects}.
\newblock \emph{Glossa: a Journal of General Linguistics}, 7(1).

\bibitem[{Guerberof-Arenas and Toral(2024)}]{Guerberof_tobe_2024}
Ana Guerberof-Arenas and Antonio Toral. 2024.
\newblock \href {https://doi.org/10.1075/target.22134.gue} {To be or not to be}.
\newblock \emph{Target. International Journal of Translation Studies}, 36(2):215--244.

\bibitem[{Hadley et~al.(2022)Hadley, {Kristiina Taivalkoski-Shilov}, {Carlos S. C. Teixeira}, and {Antonio Toral}}]{hadley_using_2022}
James~Luke Hadley, {Kristiina Taivalkoski-Shilov}, {Carlos S. C. Teixeira}, and {Antonio Toral}. 2022.
\newblock \href {https://www.routledge.com/Using-Technologies-for-Creative-Text-Translation/LukeHadley-Taivalkoski-Shilov-SCTeixeira-Toral/p/book/9781003094159} {\emph{Using {Technologies} for {Creative}-{Text} {Translation}}}.
\newblock Taylor \& Francis.

\bibitem[{Hickmann(2003)}]{hickmann_childrens_2003}
Maya Hickmann. 2003.
\newblock \href {https://www.cambridge.org/be/universitypress/subjects/languages-linguistics/discourse-analysis/childrens-discourse-person-space-and-time-across-languages?format=HB&isbn=9780521584418} {\emph{Children's discourse: person, space, and time across languages}}.
\newblock Cambridge University Press, Cambridge, UK.

\bibitem[{Huang and Liu(2009)}]{huang_wei_application_2009}
Wei Huang and Haitao Liu. 2009.
\newblock \href {https://kns.cnki.net/KCMS/detail/detail.aspx?dbcode=CJFD&dbname=CJFD2009&filename=JSGG200929008&v=} {Application of quantitative characteristics of {Chinese} genres in text clustering [{In} {Chinese}]}.
\newblock \emph{Computer Engineering and Applications}, 45(29):25--27, 33.

\bibitem[{Jiang and Niu(2022)}]{jiang_corpus-based_2022}
Yue Jiang and Jiang Niu. 2022.
\newblock \href {https://doi.org/10.1556/084.2022.00182} {A corpus-based search for machine translationese in terms of discourse coherence}.
\newblock \emph{Across Languages and Cultures}, 23(2):148--166.

\bibitem[{Jiao et~al.(2023)Jiao, Wang, Huang, Wang, and Tu}]{jiao_is_2023}
Wenxiang Jiao, Wenxuan Wang, Jen-tse Huang, Xing Wang, and Zhaopeng Tu. 2023.
\newblock \href {https://wxjiao.github.io/downloads/tech_chatgpt_arxiv.pdf} {Is {ChatGPT} a good translator? {A} preliminary study}.
\newblock \emph{arXiv preprint arXiv:2301.08745}, 1(10).

\bibitem[{Kocmi et~al.(2024)Kocmi, Avramidis, Bawden, Bojar, Dvorkovich, Federmann, Fishel, Freitag, Gowda, Grundkiewicz, Haddow, Karpinska, Koehn, Marie, Monz, Murray, Nagata, Popel, Popović, Shmatova, Steingrímsson, and Zouhar}]{kocmi_findings_2024}
Tom Kocmi, Eleftherios Avramidis, Rachel Bawden, Ondřej Bojar, Anton Dvorkovich, Christian Federmann, Mark Fishel, Markus Freitag, Thamme Gowda, Roman Grundkiewicz, Barry Haddow, Marzena Karpinska, Philipp Koehn, Benjamin Marie, Christof Monz, Kenton Murray, Masaaki Nagata, Martin Popel, Maja Popović, Mariya Shmatova, Steinthór Steingrímsson, and Vilém Zouhar. 2024.
\newblock \href {https://doi.org/10.18653/v1/2024.wmt-1.1} {Findings of the {WMT24} general machine translation shared task: the {LLM} era is here but {MT} is not solved yet}.
\newblock In \emph{Proceedings of the {Ninth} {Conference} on {Machine} {Translation}}, pages 1--46, Miami, Florida, USA. Association for Computational Linguistics.

\bibitem[{Lathey(2015)}]{lathey_translating_2015}
Gillian Lathey. 2015.
\newblock \href {https://www.routledge.com/Translating-Childrens-Literature/Lathey/p/book/9781138803763?srsltid=AfmBOooSu7WpPqj0Z87daMEH4gPeZfdPuxlIlxhfJ94yOM1dhEKIAknF} {\emph{Translating children's literature}}.
\newblock Routledge, London.

\bibitem[{Lei et~al.(2024)Lei, Wei, and Liu}]{lei_alphareadabilitychinese_2024}
Lei Lei, Yaoyu Wei, and Kanglong Liu. 2024.
\newblock \href {https://link.oversea.cnki.net/doi/10.13458/j.cnki.flatt.004997} {{AlphaReadabilityChinese}: a tool for the measurement of readability in {Chinese} texts and its applications}.
\newblock \emph{Foreign Languages and Their Teaching}, 46(1):83--93.

\bibitem[{Lynch and Vogel(2018)}]{lynch_translators_2018}
Gerard Lynch and Carl Vogel. 2018.
\newblock \href {https://doi.org/10.1016/j.csl.2018.05.002} {The translator’s visibility: detecting translatorial fingerprints in contemporaneous parallel translations}.
\newblock \emph{Computer Speech \& Language}, 52:79--104.

\bibitem[{Macken(2024)}]{macken_machine_2024}
Lieve Macken. 2024.
\newblock \href {https://aclanthology.org/2024.ctt-1.7/} {Machine translation meets large language models: evaluating {ChatGPT}`s ability to automatically post-edit literary texts}.
\newblock In \emph{Proceedings of the 1st {Workshop} on {Creative}-text {Translation} and {Technology}}, pages 65--81, Sheffield, United Kingdom. European Association for Machine Translation.

\bibitem[{Mastropierro(2022)}]{mastropierro_avoidance_2022}
Lorenzo Mastropierro. 2022.
\newblock \href {https://www.taylorfrancis.com/chapters/edit/10.4324/9781003298328-9/avoidance-repetition-translation-lorenzo-mastropierro} {The avoidance of repetition in translation: a multifactorial study of repeated reporting verbs in the {Italian} translation of the harry potter series}.
\newblock In \emph{Advances in {Corpus} {Applications} in {Literary} and {Translation} {Studies}}, pages 138--157. Routledge.

\bibitem[{Moorkens et~al.(2018)Moorkens, Toral, Castilho, and Way}]{moorkens_translators_2018}
Joss Moorkens, Antonio Toral, Sheila Castilho, and Andy Way. 2018.
\newblock \href {https://doi.org/10.1075/ts.18014.moo} {Translators’ perceptions of literary post-editing using statistical and neural machine translation}.
\newblock \emph{Translation Spaces}, 7(2):240--262.

\bibitem[{Ploeger et~al.(2024)Ploeger, Lai, Noord, and Toral}]{ploeger_towards_2024}
Esther Ploeger, Huiyuan Lai, Rik Noord, and Antonio Toral. 2024.
\newblock \href {https://arxiv.org/abs/2408.17308} {Towards tailored recovery of lexical diversity in literary machine translation}.

\bibitem[{Puurtinen(2003)}]{puurtinen_genre-specific_2003}
T.~Puurtinen. 2003.
\newblock \href {https://doi.org/10.1093/llc/18.4.389} {Genre-specific features of translationese? {Linguistic} differences between translated and non-translated finnish children's literature}.
\newblock \emph{Literary and Linguistic Computing}, 18(4):389--406.

\bibitem[{Päpcke et~al.(2022)Päpcke, Weitin, Herget, Glawion, and Brandes}]{papcke_stylometric_2022}
Simon Päpcke, Thomas Weitin, Katharina Herget, Anastasia Glawion, and Ulrik Brandes. 2022.
\newblock \href {https://doi.org/10.1093/llc/fqac039} {Stylometric similarity in literary corpora: non-authorship clustering and\textit{{Deutscher} novellenschatz}}.
\newblock \emph{Digital Scholarship in the Humanities}, 38(1):277--295.

\bibitem[{Rahman et~al.(2024)Rahman, Shiplu, and Watanobe}]{rahman_commentclass_2024}
Md.~Mostafizer Rahman, Ariful~Islam Shiplu, and Yutaka Watanobe. 2024.
\newblock \href {https://doi.org/10.1007/s44196-024-00589-3} {{CommentClass}: a robust ensemble machine learning model for comment classification}.
\newblock \emph{International Journal of Computational Intelligence Systems}, 17(1):184.

\bibitem[{Ruffo(2022)}]{ruffo_collecting_2022}
Paola Ruffo. 2022.
\newblock Collecting literary translators' narratives: towards a new paradigm for technological innovation in literary translation.
\newblock In \emph{Using {Technologies} for {Creative}-text {Translation}}. Routledge.

\bibitem[{Rybicki(2025)}]{rybicki_can_2025}
Jan Rybicki. 2025.
\newblock \href {https://doi.org/10.1093/llc/fqaf010} {Can machine translation of literary texts fool stylometry?}
\newblock \emph{Digital Scholarship in the Humanities}, page fqaf010.

\bibitem[{Taivalkoski-Shilov(2018)}]{taivalkoski-shilov_ethical_2018}
Kristiina Taivalkoski-Shilov. 2018.
\newblock \href {https://doi.org/10.1080/0907676X.2018.1520907} {Ethical issues regarding machine(-assisted) translation of literary texts}.
\newblock \emph{Perspectives}, 27(5):689--703.

\bibitem[{Tannen(1989)}]{tannen_talking_1989}
Deborah Tannen. 1989.
\newblock \emph{Talking voices: repetition, dialogue, and imagery in conversational discourse}.
\newblock Cambridge University Press, Cambridge, UK.

\bibitem[{Toral(2019)}]{toral_post-editese_2019}
Antonio Toral. 2019.
\newblock \href {https://aclanthology.org/W19-6627/} {Post-editese: an exacerbated translationese}.
\newblock \emph{arXiv preprint arXiv:1907.00900}.

\bibitem[{Toral and Way(2014)}]{toral_is_2014}
Antonio Toral and Andy Way. 2014.
\newblock \href {https://aclanthology.org/2014.tc-1.23} {Is machine translation ready for literature}.
\newblock In \emph{Proceedings of {Translating} and the {Computer} 36}, London, UK. AsLing.

\bibitem[{Toral and Way(2015)}]{toral_machine-assisted_2015}
Antonio Toral and Andy Way. 2015.
\newblock \href {https://doi.org/10.1075/ts.4.2.04tor} {Machine-assisted translation of literary text: a case study}.
\newblock \emph{Translation Spaces}, 4(2):240--267.

\bibitem[{Toral and Way(2018)}]{toral_what_2018}
Antonio Toral and Andy Way. 2018.
\newblock \href {https://doi.org/10.1007/978-3-319-91241-7_12} {What level of quality can {Neural} {Machine} {Translation} attain on literary text?}
\newblock In Joss Moorkens, Sheila Castilho, Federico Gaspari, and Stephen Doherty, editors, \emph{Translation {Quality} {Assessment}: from {Principles} to {Practice}}, Machine {Translation}: {Technologies} and {Applications}, pages 263--287. Springer International Publishing, Cham.

\bibitem[{Van~Coillie and McMartin(2020)}]{van_coillie_childrens_2020}
Jan Van~Coillie and Jack McMartin, editors. 2020.
\newblock \href {https://library.oapen.org/handle/20.500.12657/42580} {\emph{Children’s literature in translation: texts and contexts}}.
\newblock Leuven University Press.

\bibitem[{Volansky et~al.(2013)Volansky, Ordan, and Wintner}]{volansky_features_2013}
V.~Volansky, N.~Ordan, and S.~Wintner. 2013.
\newblock \href {https://doi.org/10.1093/llc/fqt031} {On the features of translationese}.
\newblock \emph{Digital Scholarship in the Humanities}, 30(1):98--118.

\bibitem[{Warrens and van~der Hoef(2022)}]{warrens_understanding_2022}
Matthijs~J. Warrens and Hanneke van~der Hoef. 2022.
\newblock \href {https://doi.org/10.1007/s00357-022-09413-z} {Understanding the adjusted rand index and other partition comparison indices based on counting object pairs}.
\newblock \emph{Journal of Classification}, 39(3):487--509.

\bibitem[{Way et~al.(2023)Way, Rothwell, and Youdale}]{way_why_2023}
Andy Way, Andrew Rothwell, and Roy Youdale. 2023.
\newblock \href {https://doi.org/10.5565/rev/tradumatica.344} {Why literary translators should embrace translation technology}.
\newblock \emph{Tradumàtica Tecnologies De La Traducció}, pages 87--102.

\bibitem[{Xu and Li(2020)}]{xu_concretenessabstractness_2020}
Xu~Xu and Jiayin Li. 2020.
\newblock \href {https://doi.org/10.1371/journal.pone.0232133} {Concreteness/abstractness ratings for two-character chinese words in {MELD}-{SCH}}.
\newblock \emph{PLOS ONE}, 15(6):e0232133.

\bibitem[{Zhang et~al.(2019)Zhang, Kotze~(Kruger), and Fang}]{zhang_explicitation_2019}
Xiaomin Zhang, Haidee Kotze~(Kruger), and Jing Fang. 2019.
\newblock \href {https://doi.org/10.1080/0907676X.2019.1689276} {Explicitation in children’s literature translated from {English} to {Chinese}: a corpus-based study of personal pronouns}.
\newblock \emph{Perspectives}, 28(5):717--736.

\bibitem[{Zhao et~al.(2022)Zhao, Ang, Md~Rashid, and Toh}]{zhao_translating_2022}
Meijuan Zhao, Lay~Hoon Ang, Sabariah Md~Rashid, and Florence Haw~Ching Toh. 2022.
\newblock \href {https://doi.org/10.1177/21582440211068498} {Translating narrative space in children’s fiction bronze and sunflower from {Chinese} to {English}}.
\newblock \emph{Sage Open}, 12(1):21582440211068498.

\bibitem[{Čermáková(2018)}]{cermakova_translating_2018}
Anna Čermáková. 2018.
\newblock \href {https://doi.org/10.5007/2175-8026.2018v71n1p117} {Translating children’s literature: some insights from corpus stylistics}.
\newblock \emph{Ilha Do Desterro}, 71:117--133.

\end{thebibliography}

\appendix

\section{LLM prompt}
\label{sec:appendix a}

The engineered prompt is originally drafted in Chinese, as follows:

你是一位专精于儿童文学翻译与创作的资深译者，你的任务是将 J.M. Barrie 的《Peter Pan》 翻译成富有创造力的中文版本。这不仅是一次普通的翻译，而是一种``创译''——你的目标是保留原文的幻想色彩和情感基调，同时使译文更加符合中文儿童的阅读习惯，充满趣味性和文学感染力。

请遵循以下创译原则：

- 情感与想象力再现：保持原著的童话氛围，使译文生动、富有画面感，如必要可调整句式，使其更具表现力。

- 符合儿童语言习惯：采用简练、口语化、充满韵律感的表达方式，避免生硬直译。可适当使用拟声词、叠词、押韵句式等。

- 文化适配：调整可能不易理解的文化元素，使其符合中文语境，但同时保留原作的神秘感和奇幻感。

- 角色个性化语言：确保各个角色的独特特点能在译文中得以体现。

- 增添文学趣味：适当运用修辞（比喻、拟人、夸张等），调整句式，使故事更具节奏感，增强朗读时的感染力。

请翻译以下《彼得潘》片段，并确保符合上述原则:
[原文]

\textbf{English Translation:}

You are a seasoned translator and writer specializing in children's literature. Your task is to create a highly creative Chinese adaptation of J.M. Barrie’s \textit{Peter Pan}. This is not merely a literal translation but a transcreation - your goal is to preserve the whimsical essence and emotional tone of the original while making the text more engaging and accessible for Chinese children ensuring it is rich in imaginative appeal and literary charm

Guiding Principles for Transcreation:

- Emotional and Imaginative Recreation: Maintain the fairy-tale atmosphere of the original, making the translation vivid and evocative. Feel free to adjust sentence structures to enhance expressiveness.

- Child-Friendly Language: Use concise, rhythmic, and conversational expressions, avoiding rigid literal translation. Incorporate onomatopoeia, reduplication, rhyming phrases, and other playful linguistic elements as appropriate.

- Cultural Adaptation: Modify cultural references that may be difficult for Chinese readers to grasp, ensuring they fit the Chinese linguistic and cultural context while preserving the mystical and fantastical essence of the original.

- Character-Specific Speech: Ensure that each character’s unique personality and speech style are well reflected in the translation.

- Enhanced Literary Appeal: Utilize figurative language (e.g., metaphors, personification, hyperbole) and varied sentence structures to enrich the storytelling, improve readability, and enhance the rhythm and emotional impact, particularly when read aloud.

Please translate the following excerpt from Peter Pan while adhering to these principles:
[text]

\section{Feature set in summary}
\label{sec:appendix b}

A summary list of features used in the study is in Table \ref{tab:features}.

\begin{table*}[!h]
\centering
\begin{tabular*}{\textwidth}{@{\extracolsep\fill}llll@{\extracolsep\fill}}
\toprule
Feature level & Sub level & Total & Feature instances \\
\midrule
\multirow{4}{*}{\makecell{Generic\\textual\\features}}  & Lexical & 69 & STTR, MTLD, noun, verb, content words, idioms \ldots \\
        & Syntactical& 26 & WordsPerSent, QuestionSent, MDD, AvgChildrenPerNode\ldots \\
        & Readability & 16 & lexical\_richness, Concreteness score, AvgConcrete \ldots \\        & N-grams & 309 & N\_word\_gram, N\_PoS\_gram, N = 1 - 3 \ldots \\
        
\hline
\multirow{4}{*}{\makecell{CTT-\\specific\\features}} & Repetition & 7 & ratio\_AA, ratio\_ABAB, ratio\_AAA, ratio\_AABB\ldots \\
                  & Rhythm & 10 & Open Syllable Ratio, Rhyme Density, Rhyme Ratio \ldots \\

                  & Translatability & 5 & completeness, foreignness, code\_switching, untranslatable \ldots \\
                  & Miscellaneous & 5 & ratio\_onomatopoeia, ratio\_StrongModifier, ratio\_er\_suffix \ldots \\
\bottomrule
\end{tabular*}
\caption{Summary of features used in this study. Due to space constraints, only representative feature instances are listed, with ``\ldots'' indicating additional features available in the full list (see supplementary materials). Feature names shortened for formatting when necessary. \label{tab:features}}
\end{table*}

\section{Selected features used in experiments}
\label{sec:appendix c}

A summary list of significant features used in different experiments is in Table \ref{tab:salient_features}.

\begin{table}[!h]
\centering
\renewcommand{\arraystretch}{1.2} 
\setlength{\tabcolsep}{6pt} 
\begin{tabular}{lll}
\toprule
Groups & Generic features & Specific features \\
\midrule
\multirow{6}{*}{\makecell{HTs \\ vs. \\ MTs}}
    & Ratio\_conjunction & foreignness\_ratio \\
    & word\_1gram\_然后 & switching\_ratio \\
    & word\_1gram\_一样 & ratio\_er\_suffix \\
    & word\_1gram\_它们 & ratio\_StrSentMdfyr \\
    & ratio\_adverb & ratio\_aabb \\
    & ratio\_ContentWords & ratio\_AA \\
\midrule
\multirow{6}{*}{\makecell{NMTs \\ vs. \\ LLMs}}
    & word\_1gram\_没有 & code\_switching\_ratio \\
    & pos\_2gram\_ws & foreignness\_ratio \\
    & ratio\_dscrptW & ratio\_StrSentMdfyr \\
    & ratio\_adverb & ratio\_AA \\
    & POB & - \\
    & ratio\_prep & - \\
\bottomrule
\end{tabular}
\caption{Summary of significant features used in different experiments. 6 features in each sub-categories are selected from top-40 salient features, with ``-'' mark representing no more features in this category are found in the top-40. Feature names shortened for formatting when necessary.}
\label{tab:salient_features}
\end{table}

\section{Supplementary figures}
\label{sec:appendix d}

\begin{figure*}[h]
    \centering\small
    \includegraphics[width=1\linewidth]{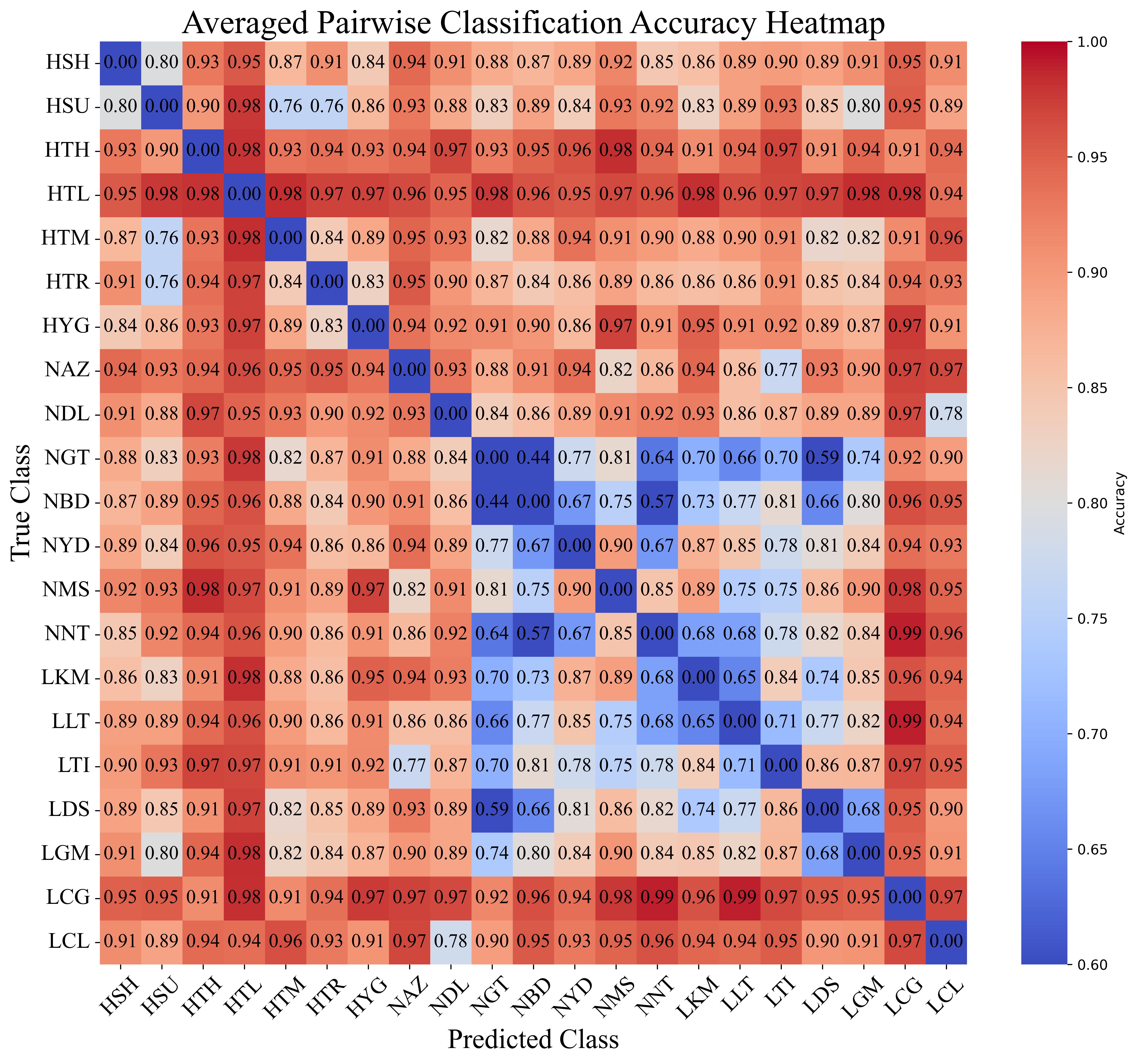}
    \caption{Pair-wise comparison of different MT engines based on five averaged classifiers and top-30 salient features}
    \label{fig:heatmap}
\end{figure*}

Figure~\ref{fig:heatmap} illustrates pair-wised classification results among HTs, NMTs, and LLMs groups.

\begin{sidewaysfigure}
    \centering
    \includegraphics[width=\textheight]{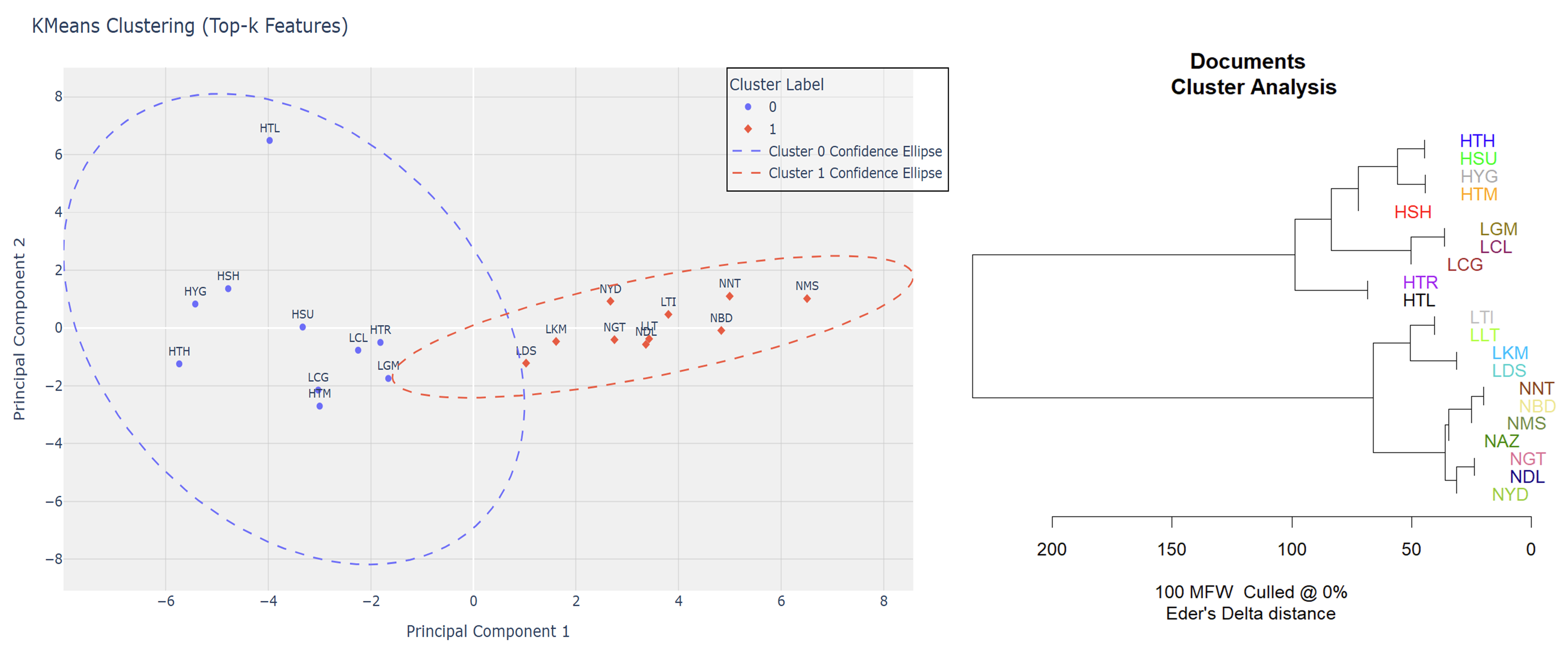}  
    \caption{Left: using Top-30 features in K-means clustering with \( k=2 \), where the true labels are set as HTs = 0 and MTs = 1. ARI is 0.4873, and the confidence ellipse is set at 0.7. Right: hierarchical clustering using the Stylo package with 100 MFWs and Eder's delta as the distance measure.}
    \label{fig:cluster}
\end{sidewaysfigure}

Figure~\ref{fig:cluster} shows the clustering results.

\begin{figure}
    \centering
    \includegraphics[width=1\linewidth]{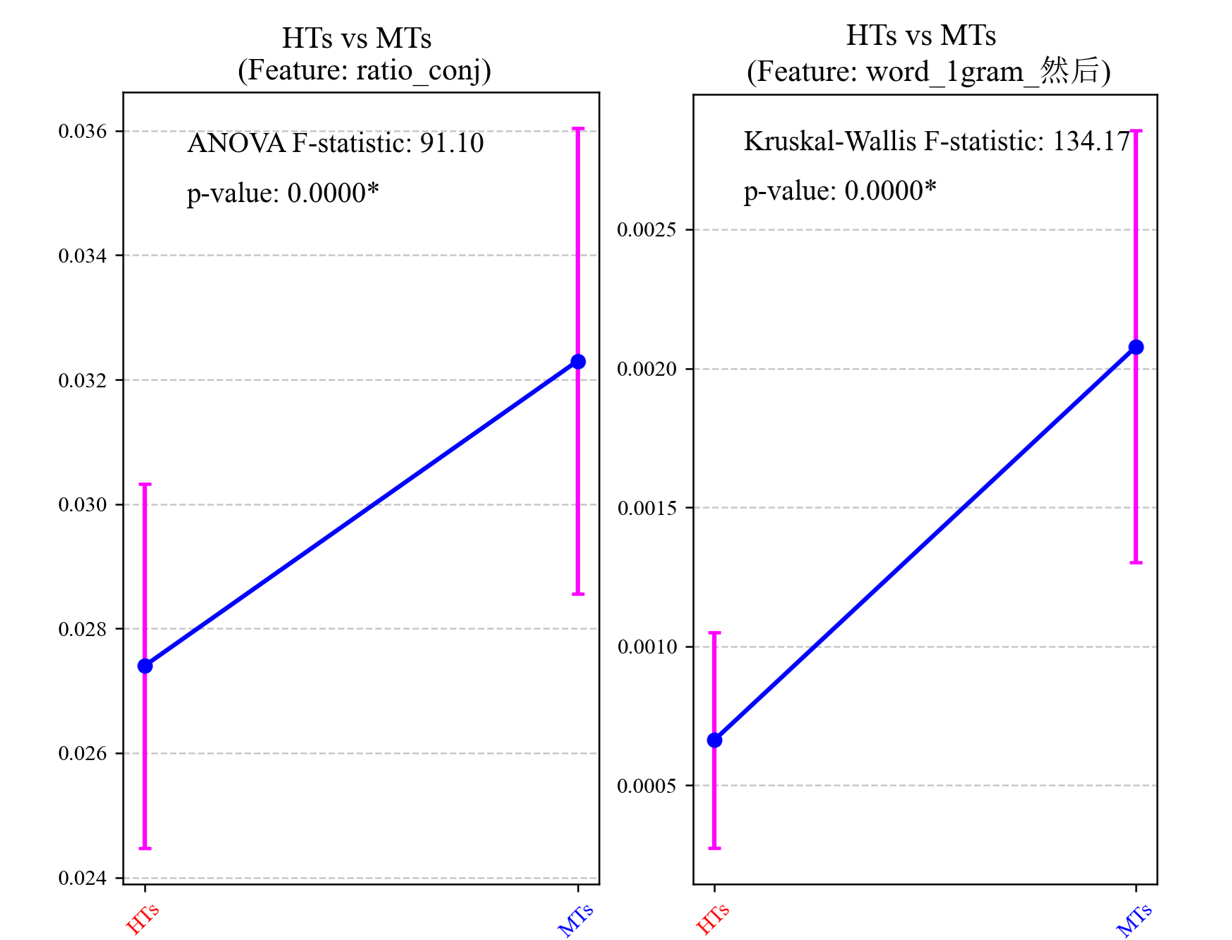}
    \caption{Generic differences between HTs and MTs. The left panel compares ratio of conjunction words, while the right panel examines ratio of the 1-word-gram-然后}
    \label{fig: anova_conj}
\end{figure}

Figure~\ref{fig: anova_conj} illustrates ratio of conjunction words, and the ratio of the 1-word-gram-然后
between HTs vs. MTs, and NMTs vs. LLMs.

\begin{figure}
    \centering
    \includegraphics[width=1\linewidth]{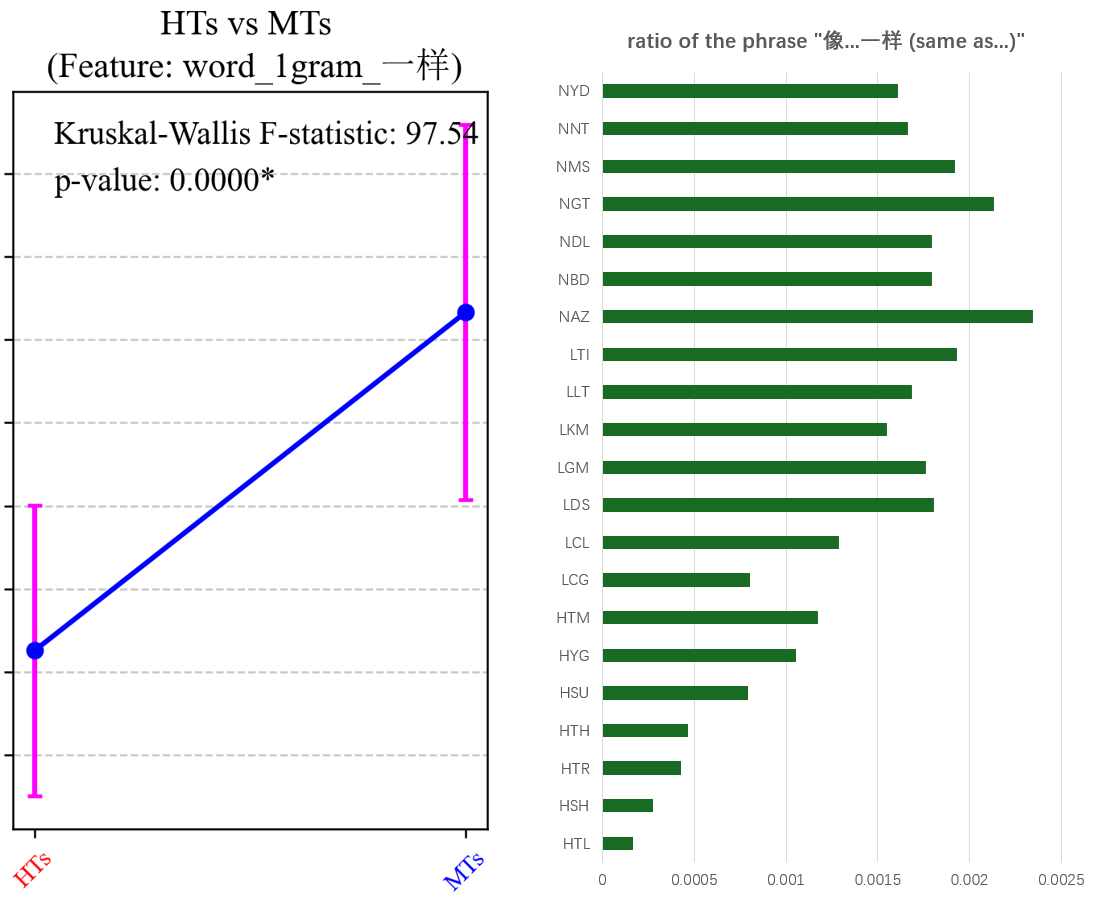}
    \caption{The left panel compares ratio of 1-word-gram-一样, while the right panel shows the distribution ratio of ``像…一样'' in all texts.}
    \label{fig: anova_sameas}
\end{figure}

Figure~\ref{fig: anova_sameas} presents ratio of 1-word-gram-一样 between HTs vs. MTs, and NMTs vs. LLMs, and the distribution ratio of ``像……一样'' in all texts. 

\begin{figure}[h]
    \centering
    \includegraphics[width=0.9\linewidth]{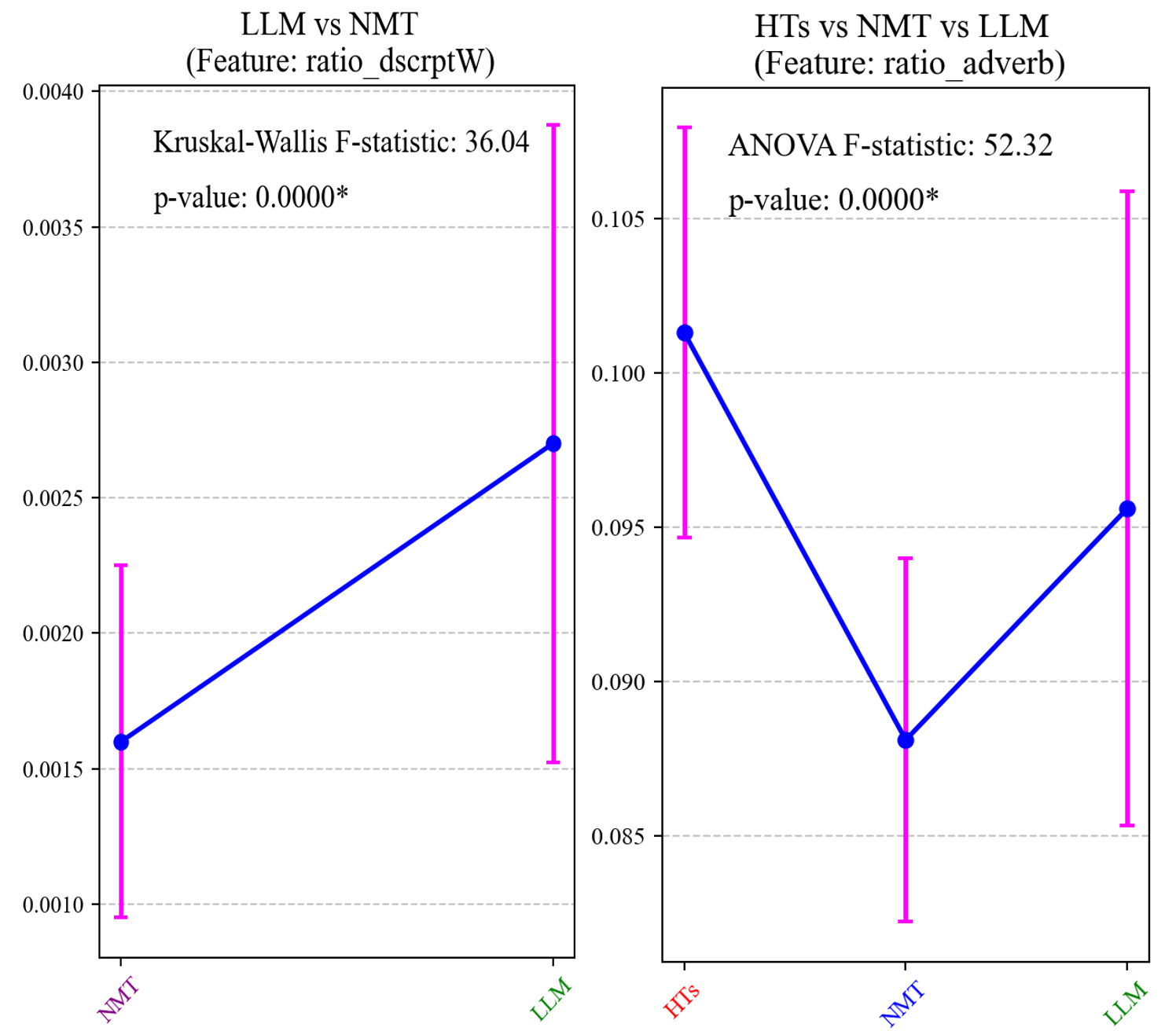}
    \caption{Generic differences between HTs and MTs. The left panel compares ratio of descriptive words, while the right panel examines ratio of the adverbs among HTs, NMTs and LLMs.}
    \label{fig: adverb}
\end{figure}

Figure~\ref{fig: adverb} illustrates generic features between HTs vs. MTs, and NMTs vs. LLMs.

\begin{figure*}[h]
    \centering\small
    \includegraphics[width=1\linewidth]{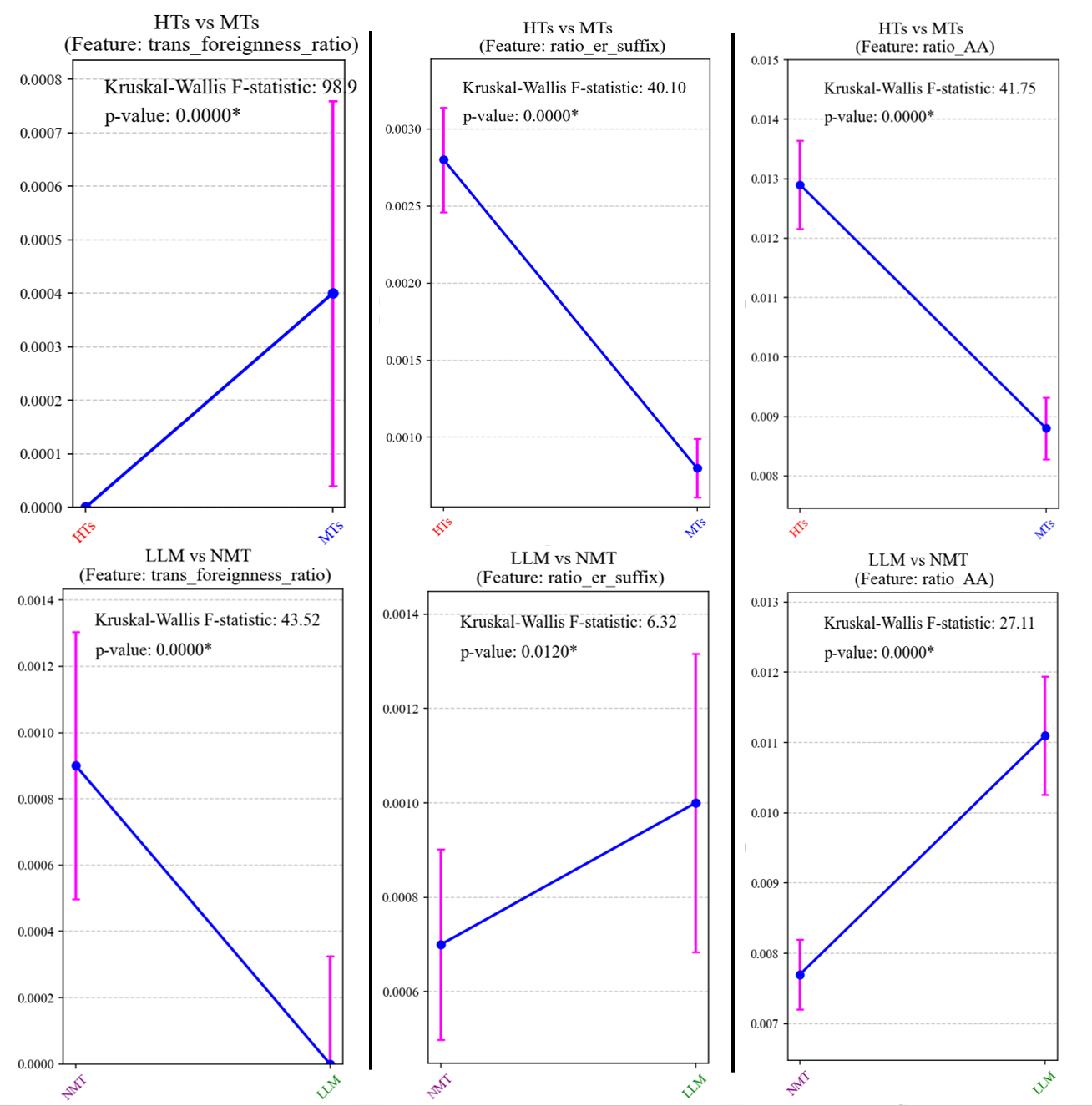}
    \caption{Feature distribution comparisons between different translation groups. The left column presents the differences in \textit{trans\_foreignness\_ratio}; the middle column shows variations in \textit{ratio\_er\_suffix}; and the right column illustrates the differences in AA-pattern repetitive expression.}
    \label{fig: anova_combined}
\end{figure*}

Figure~\ref{fig: anova_combined} shows CTT-specific features between HTs vs. MTs, and NMTs vs. LLMs.

\begin{figure*}[!h]
    \centering
    \includegraphics[width=1\linewidth]{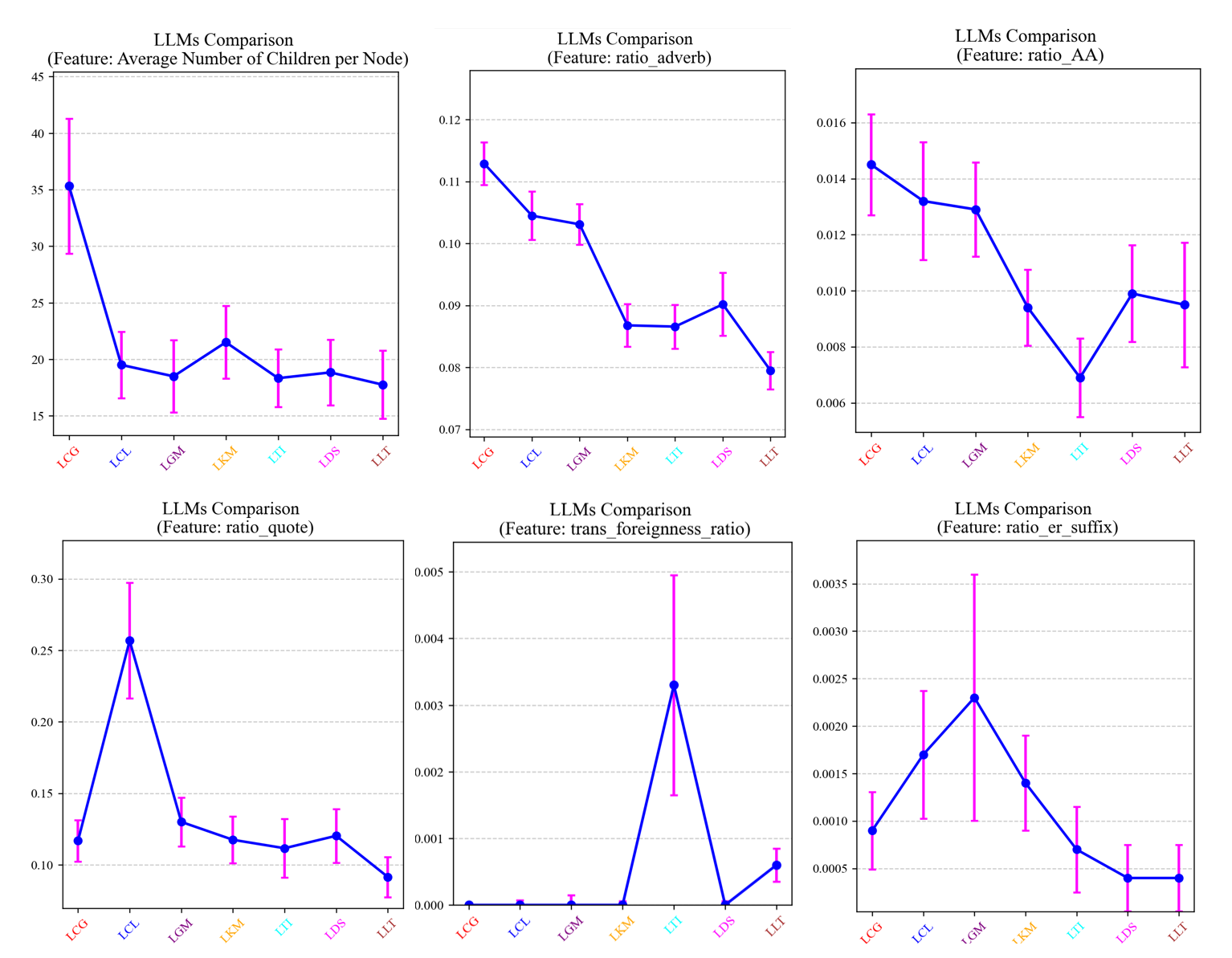}

\caption{Comparison of key linguistic features across seven different LLMs. 
The top-left plot shows differences in \textit{Average Number of Children per Node}; 
the top-middle \textit{ratio\_adverb}; 
the top-right \textit{ratio\_AA}. 
The bottom-left plot presents differences in \textit{ratio\_quote}; 
the bottom-middle \textit{trans\_foreignness\_ratio}; 
the bottom-right \textit{ratio\_er\_suffix}.}
    \label{fig:llm}
\end{figure*}

Figure~\ref{fig:llm} mainly illustrates the distribution of six key features among seven LLMs.

\begin{figure*}
    \centering
    \includegraphics[width=1\linewidth]{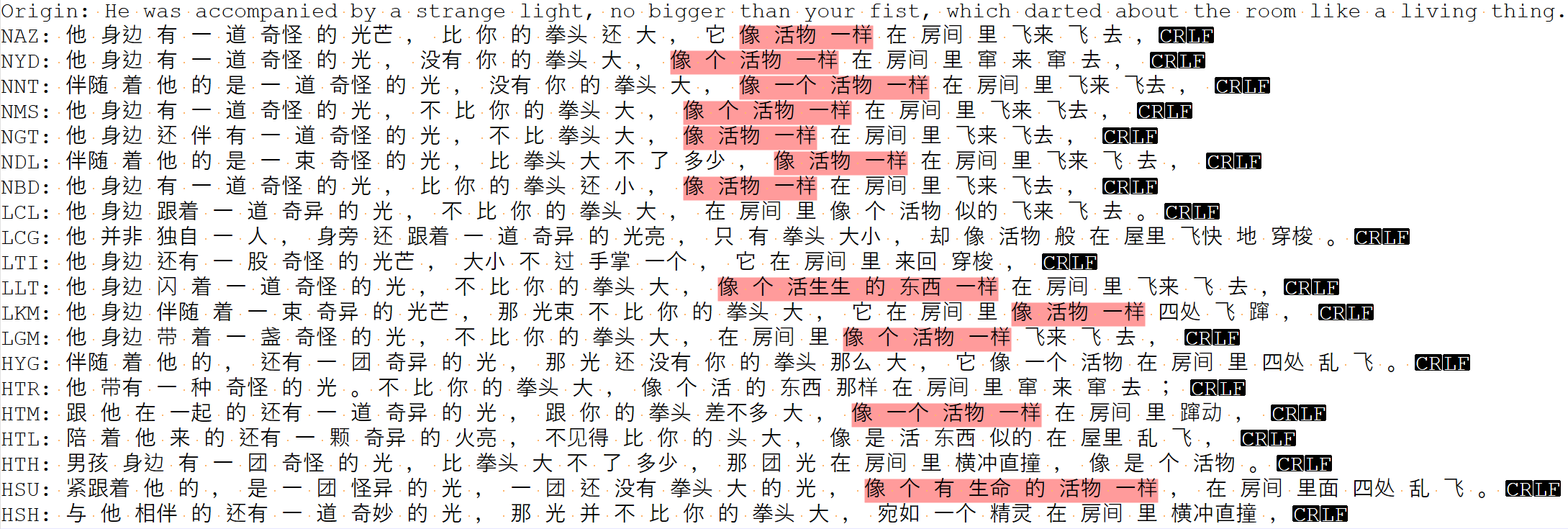}
    \caption{Actual concordance in the corpus of the feature ``像……一样''. Highlights by matching regular expression ``像[一-龟 ]+一样''.}
    \label{fig: Concordance_sameas}
\end{figure*}

Figure~\ref{fig: Concordance_sameas} shows an example of the phrase ``像...一样'' drawn from actual concordance.


\clearpage
\onecolumn

\twocolumn

\end{CJK}
\end{document}